\PassOptionsToPackage{unicode}{hyperref}
\PassOptionsToPackage{hyphens}{url}
\PassOptionsToPackage{dvipsnames,svgnames,x11names}{xcolor}
\documentclass[
]{scrartcl}

\usepackage{amsmath,amssymb}
\usepackage{iftex}
\ifPDFTeX
  \usepackage[T1]{fontenc}
  \usepackage[utf8]{inputenc}
  \usepackage{textcomp} 
\else 
  \usepackage{unicode-math}
  \defaultfontfeatures{Scale=MatchLowercase}
  \defaultfontfeatures[\rmfamily]{Ligatures=TeX,Scale=1}
\fi
\usepackage{lmodern}
\ifPDFTeX\else  
\fi
\IfFileExists{upquote.sty}{\usepackage{upquote}}{}
\IfFileExists{microtype.sty}{
  \usepackage[]{microtype}
  \UseMicrotypeSet[protrusion]{basicmath} 
}{}
\makeatletter
\@ifundefined{KOMAClassName}{
  \IfFileExists{parskip.sty}{%
    \usepackage{parskip}
  }{
    \setlength{\parindent}{0pt}
    \setlength{\parskip}{6pt plus 2pt minus 1pt}}
}{
  \KOMAoptions{parskip=half}}
\makeatother
\usepackage{xcolor}
\ifx\paragraph\undefined\else
  \let\oldparagraph\paragraph
  \renewcommand{\paragraph}[1]{\oldparagraph{#1}\mbox{}}
\fi
\ifx\subparagraph\undefined\else
  \let\oldsubparagraph\subparagraph
  \renewcommand{\subparagraph}[1]{\oldsubparagraph{#1}\mbox{}}
\fi

\providecommand{\tightlist}{%
  \setlength{\itemsep}{0pt}\setlength{\parskip}{0pt}}\usepackage{longtable,booktabs,array}
\usepackage{calc} 
\usepackage{etoolbox}
\makeatletter
\patchcmd\longtable{\par}{\if@noskipsec\mbox{}\fi\par}{}{}
\makeatother
\IfFileExists{footnotehyper.sty}{\usepackage{footnotehyper}}{\usepackage{footnote}}
\makesavenoteenv{longtable}
\usepackage{graphicx}
\makeatletter
\def\maxwidth{\ifdim\Gin@nat@width>\linewidth\linewidth\else\Gin@nat@width\fi}
\def\maxheight{\ifdim\Gin@nat@height>\textheight\textheight\else\Gin@nat@height\fi}
\makeatother
\setkeys{Gin}{width=\maxwidth,height=\maxheight,keepaspectratio}
\makeatletter
\def\fps@figure{htbp}
\makeatother
\newlength{\cslhangindent}
\newlength{\csllabelwidth}
\newlength{\cslentryspacingunit} 
\newenvironment{CSLReferences}[2] 
 {
  \setlength{\parindent}{0pt}
  \ifodd #1
  \let\oldpar\par
  \def\par{\hangindent=\cslhangindent\oldpar}
  \fi
  \setlength{\parskip}{#2\cslentryspacingunit}
 }%
 {}
\usepackage{calc}

\usepackage[noblocks]{authblk}

\makeatletter
\@ifpackageloaded{caption}{}{\usepackage{caption}}
\AtBeginDocument{%
\ifdefined\contentsname
  \renewcommand*\contentsname{Table of contents}
\else
  \newcommand\contentsname{Table of contents}
\fi
\ifdefined\listfigurename
  \renewcommand*\listfigurename{List of Figures}
\else
  \newcommand\listfigurename{List of Figures}
\fi
\ifdefined\listtablename
  \renewcommand*\listtablename{List of Tables}
\else
  \newcommand\listtablename{List of Tables}
\fi
\ifdefined\figurename
  \renewcommand*\figurename{Figure}
\else
  \newcommand\figurename{Figure}
\fi
\ifdefined\tablename
  \renewcommand*\tablename{Table}
\else
  \newcommand\tablename{Table}
\fi
}
\@ifpackageloaded{float}{}{\usepackage{float}}
\floatstyle{ruled}
\@ifundefined{c@chapter}{\newfloat{codelisting}{h}{lop}}{\newfloat{codelisting}{h}{lop}[chapter]}
\floatname{codelisting}{Listing}

\makeatother
\makeatletter
\@ifpackageloaded{caption}{}{\usepackage{caption}}
\@ifpackageloaded{subcaption}{}{\usepackage{subcaption}}
\makeatother
\makeatletter
\@ifpackageloaded{tcolorbox}{}{\usepackage[skins,breakable]{tcolorbox}}
\makeatother
\makeatletter
\@ifundefined{shadecolor}{\definecolor{shadecolor}{rgb}{.97, .97, .97}}
\makeatother
\ifLuaTeX
  \usepackage{selnolig}  
\fi
\IfFileExists{bookmark.sty}{\usepackage{bookmark}}{\usepackage{hyperref}}
\IfFileExists{xurl.sty}{\usepackage{xurl}}{} 
\hypersetup{
  pdftitle={Collaborative Grid Mapping for Moving Object Tracking Evaluation},
  pdfauthor={Rémy Huet; Antoine Lima; Philippe Xu; Véronique Cherfaoui; Philippe Bonnifait},
  colorlinks=true,
  linkcolor={blue},
  filecolor={Maroon},
  citecolor={Blue},
  urlcolor={Blue},
  pdfcreator={LaTeX via pandoc}}

\title{Collaborative Grid Mapping for Moving Object Tracking Evaluation}

\author{Rémy Huet}
\author{Antoine Lima}
\author{Philippe Xu}
\author{Véronique Cherfaoui}
\author{Philippe Bonnifait}

\affil[1]{Heudiasyc UMR CNRS 7253, Université de technologie de
Compiègne}

\date{2023-07-24}
\begin{document}
\maketitle
\begin{abstract}
Perception of other road users is a crucial task for intelligent
vehicles. Perception systems can use on-board sensors only or be in
cooperation with other vehicles or with roadside units. In any case, the
performance of perception systems has to be evaluated against
ground-truth data, which is a particularly tedious task and requires
numerous manual operations. In this article, we propose a novel
semi-automatic method for pseudo ground-truth estimation. The principle
consists in carrying out experiments with several vehicles equipped with
LiDAR sensors and with fixed perception systems located at the roadside
in order to collaboratively build reference dynamic data. The method is
based on grid mapping and in particular on the elaboration of a
background map that holds relevant information that remains valid during
a whole dataset sequence. Data from all agents is converted in
time-stamped \textit{observations} grids. A data fusion method that
manages uncertainties combines the background map with observations to
produce dynamic reference information at each instant. Several datasets
have been acquired with three experimental vehicles and a roadside unit.
An evaluation of this method is finally provided in comparison to a
handmade ground truth.
\end{abstract}
\ifdefined\Shaded\renewenvironment{Shaded}{\begin{tcolorbox}[borderline west={3pt}{0pt}{shadecolor}, enhanced, breakable, interior hidden, boxrule=0pt, frame hidden, sharp corners]}{\end{tcolorbox}}\fi

\begin{figure}

{\centering \includegraphics{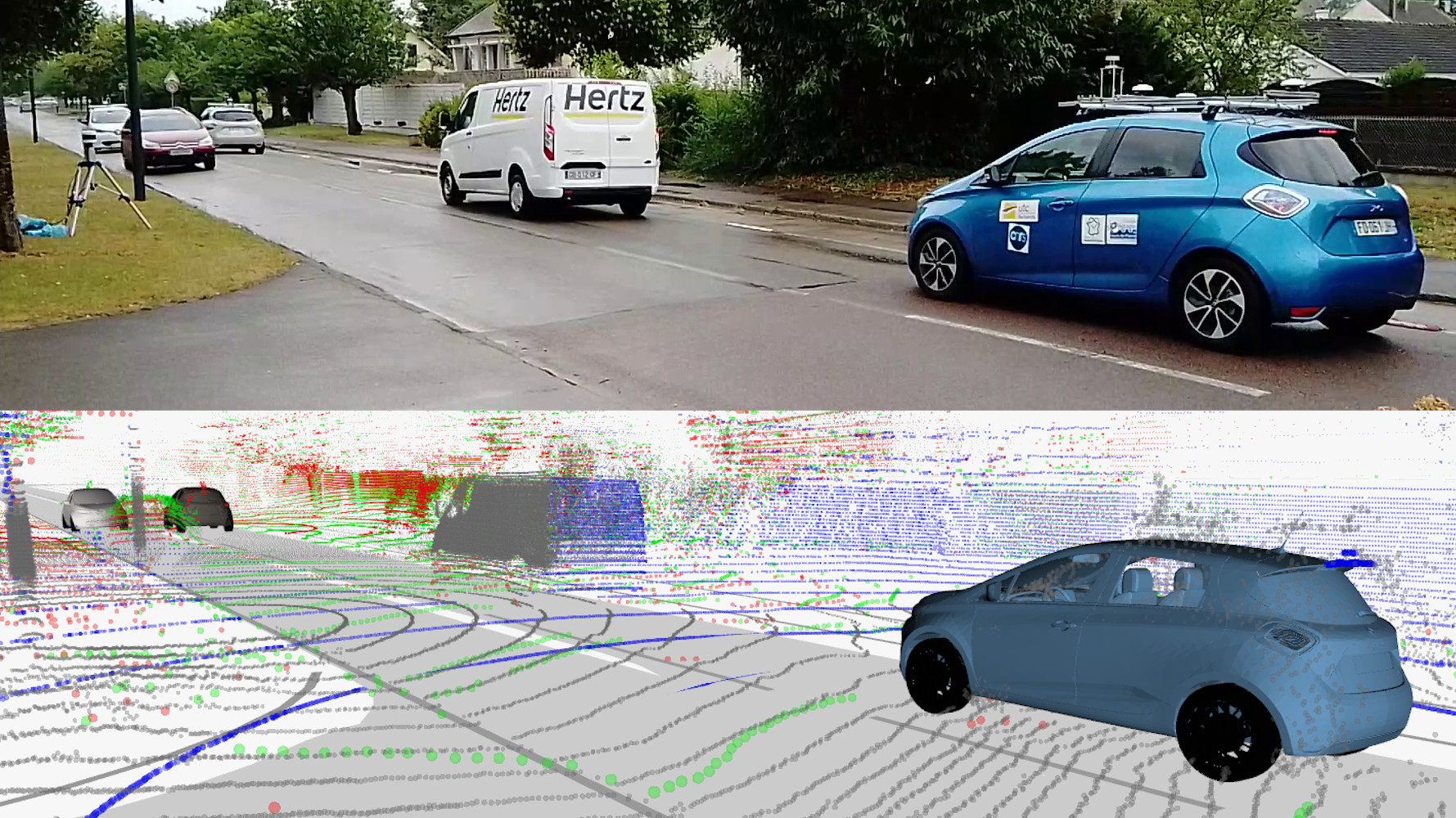}

}

\caption{\label{fig-photo_dataset}LiDAR and GNSS data gathered by
several moving vehicles and a roadside sensor before processing.}

\end{figure}

\hypertarget{sec-introduction}{%
\section{Introduction}\label{sec-introduction}}

In order to navigate safely, intelligent vehicles must perceive their
environment. Perception systems have become more and more complex and
evaluating them under real conditions is a difficult task that generally
requires manual annotation and processing. In addition to being a
time-consuming task, ground-truth annotation of perception systems is
error-prone, as occlusions can limit perceivable objects. Therefore,
this task can be especially challenging in dynamic and cluttered
environments. One way to address this issue is to use and record
complementary points of view from remote sources of information that can
be other road users or roadside units. The objective is then to merge in
a post-processing stage all the available data through space and time
into a coherent reference.

In this paper, we propose a method that builds occupancy grid maps on
recorded driving sequences for the evaluation of perception and tracking
systems using several sensors that are either embedded on vehicles or
fixed in the environment. Each agent participating in the mapping
registers the time-stamped data of its sensors, the synchronization of
the clocks being provided by Global Navigation Satellite System (GNSS)
receivers.

To represent and fuse data from various perception systems in a common
frame, bird eye view grids can be used. As such, our method provides
both a grid of the background containing all static elements and
time-stamped perception grids in which moving road users are detected.
Such grids are a first step towards the generation of a reliable ground
truth for object tracking evaluation built semi-automatically.

Moreover, a dataset is released with real data coming from three
vehicles and a road-side sensor, including three short scenarios of
interest involving a dozen road-users. To our knowledge, it is the first
dataset with real data and multiple moving points of view for
collaborative perception. A video illustrating the dataset recording and
the method is given\footnote{\url{https://hal.science/hal-04172876}}.

The article is organized as follows. Section~\ref{sec-related} gives an
overview of related work on post-processing for object tracking and on
occupancy grids. Section~\ref{sec-definition} introduces the notion of
\emph{background map} and its representation as an evidential occupancy
grid. In section Section~\ref{sec-construction}, we provide examples
about how to build this map with high-accuracy GNSS receivers, on-car
and roadside LiDAR sensors. In Section~\ref{sec-final_map}, we explain
how we construct the final grids and how the background map we created
helps to take into account past and future observations. Finally section
Section~\ref{sec-experiments} presents experimental results and
evaluation of our method on a multi-vehicle dataset recorded for this
purpose.

\hypertarget{sec-related}{%
\section{Related Work}\label{sec-related}}

Perceiving other road users is a complex but mandatory task for
autonomous vehicles. It allows the navigation system to find a safe
trajectory in the complex environment of the roads. Road users can be
represented and tracked object-wise (Vo et al. 2015) but a dense mapping
approach is particularly adapted to planning and reactive control in
opened and uncontrolled environments (Ziegler and Stiller 2010).
Introduced in (Elfes 1989), 2D probabilistic occupancy grids, that
densely represent where the space is free or occupied, have widely
gained traction in the perception community. One of their main drawback
is the difficulty to represent movement of the environment, a limitation
alleviated in (Coué et al. 2006) by using a 4D grid to represent the
position and velocity of obstacles with probabilities. Another common
extension is to use the theory of Dempster-Shafer (or theory of
evidence) to represent multiple possible states of a cell (Moras,
Cherfaoui, and Bonnifait 2011). Combining both tracking and multi-state
is referred to as Grid-Based Tracking and Mapping (GTAM) (Steyer,
Tanzmeister, and Wollherr 2018; Tanzmeister and Wollherr 2017). These
approaches often use a particle filter to estimate the motion of
obstacles in the grid.

Another important aspect of perception systems is their evaluation.
There exist many single Point of View (PoV) datasets such as KITTI
(Geiger, Lenz, and Urtasun 2012), SemanticKITTI (Behley et al. 2019) or
NuScenes (Caesar et al. 2020) for the evaluation of automotive
perception systems. They provide manually labeled datasets with
delimited 3D bounding boxes. To extend evaluations to navigation,
Ettinger et al.~(Ettinger et al. 2021) provide a perception dataset with
a focus on motion prediction. However, multi-PoV datasets are not yet
widely developed with most being based either on simulated data (Xu et
al. 2022; Yiming Li et al. 2022; Mao et al. 2022) or static sensors
(Busch et al. 2022). Recently, the DAIR-V2X dataset~(H. Yu et al. 2022)
has been released with several points of view from real sensors.
However, it is focused on vehicle-to-infrastructure communication and
only contains one vehicle. A possible reason for this is the complexity
of labeling a multi-PoV dataset, which is why several methods have been
proposed to label them semi-automatically (Han et al. 2023).

Approaches for ground-truth generation are done offline on recorded
datasets. They can take advantage from heavy-computational tasks such as
multiple point clouds registration (Ye, Spiegel, and Althoff 2020) or
the use of non-causal (i.e.~depending on future information) algorithms
(Erik Stellet, Walkling, and Marius Zöllner 2016; B. Yu and Ye 2020). In
(B. Yu and Ye 2020), the tracks corresponding to other road users are
created at the most certain point in the dataset before being propagated
forward and backward by filtering and smoothing algorithms. Other
approaches (Ye et al. 2021) are based on GTAM techniques and use offline
post-processing to make a backward smoothing pass on the particle filter
to help its convergence.

Our method builds on previously cited GTAM techniques (Steyer,
Tanzmeister, and Wollherr 2018; Tanzmeister and Wollherr 2017; Ye et al.
2021). A finer frame of discernment allows us to better represent the
environment to take advantage of segmentation algorithms to gather more
information. We use the possibilities offered by offline post-processing
to gather all the information of a recording in a ``background map'',
that is then fused with observation grids.

\hypertarget{sec-definition}{%
\section{Definition of the Background Map}\label{sec-definition}}

\hypertarget{sec-map_def}{%
\subsection{Semantics of the Map}\label{sec-map_def}}

In general, the mapping of an environment holds information that is true
over very long periods of time. Here, the information stored by the
mapping is defined as the information that is reliable during the whole
dataset. It represents statically occupied areas and passable areas,
which can be either free or contain a moving object.

Occupancy of the space can be divided into two main classes:
\emph{immovable occupancy} that corresponds to non-movable features
(e.g.~buildings, trees or road signals), and \emph{objects} that
corresponds to movable features (i.e.~vehicles or vulnerable road
users). The distinction between those two classes can be made in a
single measurement, while objects have to be refined into \emph{static}
or \emph{dynamic} using multiple measurements. As the goal of the map is
to determine some information that is valid for the whole dataset,
static objects are defined as objects that do not move during the whole
sequence. Dynamic objects are thus objects that moved at least once
during the sequence.

\hypertarget{sec-evidential_representation}{%
\subsection{Evidential Grid
Representation}\label{sec-evidential_representation}}

The map has to represent several classes and combinations, a problem
well suited for the Dempster-Shafer Theory (DST) framework (Dempster
1968; Shafer 1976). In the DST, a \emph{mass function} \(m\) assigns a
non-negative value to each element of a power set \(2^\Theta\) where
\(\Theta\) is the \emph{frame of discernment}, the exhaustive and atomic
set of hypotheses about a variable, such that
\(\sum_{A\in 2^\Theta} m(A) = 1\).

This framework allows a fine representation of uncertainty by assigning
some mass to a set of classes, including the whole frame of discernment
to represent full uncertainty. This is especially suited to the case of
partial sensor information, which can be represented by a union of
classes but not as an atomic hypothesis from the frame. Mass functions
can also be combined using for example the conjunctive rule:
\begin{equation}\protect\hypertarget{eq-dempster}{}{
m_{1, 2}(A) = \sum_{B\cap C = A} m_1(B)\cdot m_2(C),~\forall A\in 2^\Theta
}\label{eq-dempster}\end{equation}

To represent this information, we project the 3D data into a 2D discrete
Grid Map at the ground level. 2D grids provide an efficient framework
for data representation for terrestrial vehicles and help for data
fusion in a common spatial frame.

\hypertarget{sec-fod}{%
\subsection{Frame of Discernment}\label{sec-fod}}

The frame of discernment to characterize an environment is defined as
follows: \begin{equation}\protect\hypertarget{eq-frame_discernment}{}{
    \Theta = \left\{F, I, S, D\right\},
}\label{eq-frame_discernment}\end{equation} where \(F\) stands for free
space, \(I\) for immovable, \(S\) for static objects and \(D\) for
dynamic objects, as defined in Section~\ref{sec-map_def}. For
convenience, we add the following definitions:

\begin{itemize}
\tightlist
\item
  \(M = \left\{S, D\right\}\) for movable objects that can be static or
  dynamic;
\item
  \(O = \left\{I, S, D\right\}\) for generic unclassified occupancy;
\item
  \(P = \left\{F, D\right\}\) which stands for \emph{Passable} as
  defined in (Steyer, Tanzmeister, and Wollherr 2018), corresponding to
  zones that are either free or occupied by a dynamic object.
\end{itemize}

Finally, cells of the grid without any information will have all the
mass assigned to \(\Theta\).

As the map only contains information that is valid during the whole
dataset, its frame of discernment does not include time-dependent
classes such as dynamic objects and free space. It is a 2D spatial
discrete structure, in which each cell contains a mass function defined
on the following frame of discernment:

\begin{equation}\protect\hypertarget{eq-map-frame}{}{
\Theta_{\mathcal{M}} = \left\{ P, I, S\right\}
}\label{eq-map-frame}\end{equation}

Indeed, information about passable areas, immovable occupancy and static
objects is time independent during a whole sequence, contrary to dynamic
objects and free space that should not appear in the background map.

Compared to the frame of discernment used by previous approaches
(Steyer, Tanzmeister, and Wollherr 2018; Tanzmeister and Wollherr 2017),
ours allows a finer representation thanks to the distinction between
\emph{immobile} and \emph{mobile} features. This distinction allows us
to integrate class information from segmentation algorithms, as
explained in the next section.

\hypertarget{sec-construction}{%
\section{Background Map Construction}\label{sec-construction}}

\begin{figure}

{\centering \includegraphics{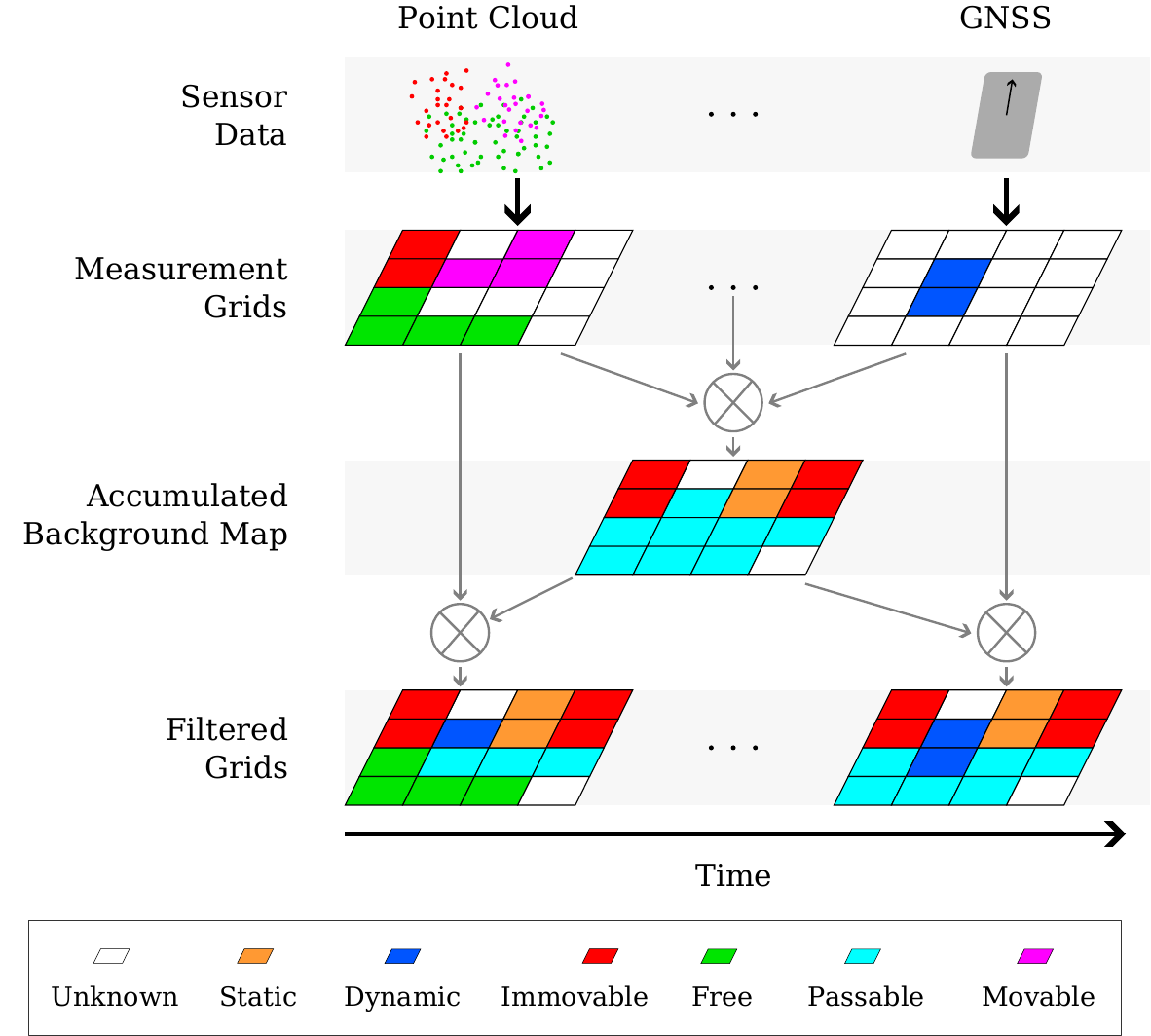}

}

\caption{\label{fig-mapping}Schematic presentation of the method. Sensor
data is turned into \emph{Measurement Grids}. These grids are then fused
into a \emph{Background Map}. Finally, \emph{Measurement Grids} are
filtered through time and fused with the \emph{Background Map} to
generate \emph{Filtered Grids} at each time step.}

\end{figure}

\hypertarget{sec-construction:map_generation}{%
\subsection{Collaborative Background Map
Generation}\label{sec-construction:map_generation}}

The method to construct the background map takes different sources
providing information about obstacles or free space in a common
East-North-Up frame. GNSS and LiDAR sensors are used in this work, but
other sensors could be adapted such as cameras or RADARs. The background
map is generated in several steps, as illustrated in
Figure~\ref{fig-mapping}.

\hypertarget{observation-grid-generation}{%
\subsubsection{Observation Grid
Generation}\label{observation-grid-generation}}

As illustrated in Figure~\ref{fig-mapping}, observations are grids
constructed from all sources separately at different instants by
projecting sensor data in 2D. Depending on the sensor, different
evidences can be derived such as free space \(F\), occupancy \(O\) or
dynamic objects \(D\).

\hypertarget{map-size-deduction}{%
\subsubsection{Map Size Deduction}\label{map-size-deduction}}

In addition to being stored for further processing, observations are
first used to determine the size of the background map as being wide
enough to fuse all the observations.

\hypertarget{data-accumulation}{%
\subsubsection{Data accumulation}\label{data-accumulation}}

The map is then constructed in a second pass by accumulating the
previously stored observations, as shown in Figure~\ref{fig-mapping}.
Successive observations are accumulated on a per-cell basis. Each cell
\(c\) stores the values of its consecutive observations \(m_c^t\)
through time \(t\) in an observation buffer
\(\mathbf{m}_c = \left\{\dots, m_c^t, \dots\right\}\). The longest
string of mostly free, mostly immovable and mostly static beliefs in
\(\mathbf{m}_c\) are then computed as \(w_f\), \(w_i\) and \(w_s\),
respectively, to determine the overall state of cell \(c\). For this,
two thresholds \(t_f\) and \(t_o\) are defined as the minimal number of
\emph{consecutive} free and occupancy observations required to consider
that a cell is free or occupied in a noise resilient manner:
\begin{equation}\protect\hypertarget{eq-cell-classification}{}{
s_c =
    \begin{cases}
        P & \text{if } |w_f| \geq t_f;\\
        I & \text{else if } |w_i + w_s| \geq t_o \text{ and } |w_i| \geq |w_s|;\\
        S & \text{else if } |w_i + w_s| \geq t_o \text{ and } |w_s| > |w_i|;\\
        \Theta & \text{otherwise}
    \end{cases}
}\label{eq-cell-classification}\end{equation}

\begin{figure}

\begin{minipage}[b]{0.50\linewidth}

{\centering 

\includegraphics{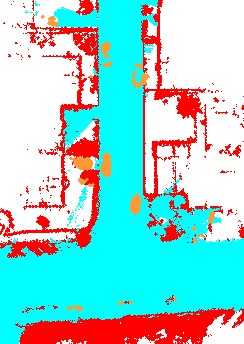}

}

\subcaption{\label{fig-thresholds-low}}
\end{minipage}%
\begin{minipage}[b]{0.50\linewidth}

{\centering 

\includegraphics{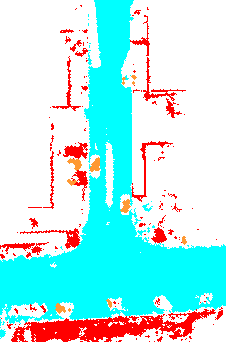}

}

\subcaption{\label{fig-thresholds-high}}
\end{minipage}%

\caption{\label{fig-thresholds}Effects of threshold tuning for the map
generation. In (a), free space threshold is too low, leading noise to
create passable areas inside vehicles (cyan regions inside orange
borders). In (b), free space and occupancy thresholds are too high,
leading to a complete lack of information and the creation of unknown
zones (white).}

\end{figure}

The thresholds are a trade-off between the completeness of the map and
its coherence. Augmenting them will result in more unknown areas, while
decreasing them could lead to unwanted passable or occupied evidence. To
tune the thresholds, we ensure that the static and immovable obstacles
are well mapped, and that there is no ghost (cells classified as static
object because an object passed on them for too long) in the map.
Figure~\ref{fig-thresholds} shows the effect of tuning the thresholds in
our experiments. We keep them as low as possible while being consistent
to avoid the creation of unknown zones in the map.

\hypertarget{segmentation-of-lidar-point-clouds}{%
\subsection{Segmentation of LiDAR Point
Clouds}\label{segmentation-of-lidar-point-clouds}}

In the current implementation of the method, we use point clouds from
rotating LiDARs on-board vehicles or installed on fixed supports at the
roadside. This section describes the processing applied to point clouds
to extract information used in the generation of the map.

Raw point cloud data in itself is insufficient to derive the pieces of
evidence required for the map generation process. However, it is
processed to extract such evidence. For example, points hitting the
ground support the free space hypothesis while points hitting a building
support the infrastructure hypothesis. Such processing is described
hereafter for each source and time step.

\hypertarget{ground-segmentation}{%
\subsubsection{Ground Segmentation}\label{ground-segmentation}}

A common task on point clouds is to segment non-ground points from
ground ones. Such segmentation allows to make the difference between
points on the ground supporting the free space evidence \(F\) from
points above the ground, supporting the existence of an obstacle and
thus some unclassified occupancy \(O\).

For this purpose, we use the method from (Jiménez et al. 2021) which
gives very good results with an F1-score above \(95\%\) for obstacle
detection on nuScenes dataset (Caesar et al. 2020). This method is based
on gradient and height difference between consecutive points in a
vertical scan of the LiDAR. Ground height is then estimated in a
cylindrical voxel using loopy belief propagation to refine the
segmentation previously done.

\hypertarget{class-segmentation}{%
\subsubsection{Class Segmentation}\label{class-segmentation}}

In parallel, points are classified to refine the unclassified occupancy
\(O\). This task (called class segmentation) has seen major improvements
in the previous years, in particular based on neural networks (Zamanakos
et al. 2021; Ying Li et al. 2020). We use the method from (Zhu et al.
2021) which achieved an \emph{Intersection over Union} (IoU) score above
\(93\%\) for car segmentation in both nuScenes and SemanticKITTI (Behley
et al. 2019) datasets. This method is based on asymmetrical convolution
networks on cylindrical voxels, that match to the polar nature of a
rotating LiDAR. A per-point refinement is applied to segment the point
cloud. The overall \emph{mean IoU} (mIoU) is around \(68 \%\) on
SemanticKITTI and \(76 \%\) on nuScenes. Other approaches such as (Chen
et al. 2021; Dewan et al. 2016) propose instead to segment which point
is moving across time or not, a task called \emph{Moving-Object
Segmentation}. Such a technique could be used to derive dynamic
evidence, but would not provide a class for obstacles. Although class
information is not directly used for ground-truth generation, it is a
precious semantic information that we would like to keep.

Class segmentation thus provides evidence based on class, with for
example the ground class supporting free space \(F\), non-movable
classes (i.e.~buildings, fences or vegetation) supporting Infrastructure
\(I\) and movable classes (i.e.~cars, trucks or pedestrians) supporting
movable \(M\). As, the performance of the chosen deep learning method is
not high enough with our LiDARs, the segmented class has only been used
to refine points labeled as obstacles by the ground segmentation of
(Jiménez et al. 2021).

\hypertarget{sec-final_map}{%
\section{Final Map for Tracking Evaluation}\label{sec-final_map}}

Using the previously generated background map and observation grids, we
compute ground-truth maps that characterize free space and dynamic
objects (see Figure~\ref{fig-mapping}). These maps are obtained at each
time step by fusing the background map and the observations with the
conjunctive rule. This fusion allows to take into account information
from the whole dataset given by the background map at each timestamp. As
is, occupancy \(O\) seen at a time \(t\) in a passable area \(P\) will
lead to the creation of dynamic cells \(D\), as \(O \cap P = D\). The
observation of occupancy in immovable or static zone will result in the
same immovable or static evidence, and free space will be added on top
of passable zones when observed. In the case of incoherent information
between the background map and the observation, we use the information
from the map as an output as it results from information from the whole
dataset and is less subject to noise than instantaneous observations.

The final result provided by the method is then a 2D evidential grid
containing all classes defined in
Section~\ref{sec-evidential_representation}. Passable zones resulting
from this fusion correspond to non-observed zones at this timestamp, and
are just some information provided by the map. These grids can be used
for moving object tracking evaluation as they contain cells occupied by
dynamic objects. Static objects are contained in the map and their
detection can then be evaluated, or one may choose to ignore them in the
validation process if it is irrelevant.

\hypertarget{sec-experiments}{%
\section{Experiments}\label{sec-experiments}}

\hypertarget{sec-dataset}{%
\subsection{Collaborative Dataset}\label{sec-dataset}}

As highlighted in the review of related work, to our knowledge, no
dataset provides all at once moving, multi-vehicle, with real data. To
develop and test our method, we thus recorded a multi-vehicle dataset
which is available online\footnote{\url{https://datasets.hds.utc.fr/share/KrXdEBzaMnMmWbV}}.
This dataset contains three scenarios involving three vehicles equipped
with sensors and a roadside sensor. These scenarios have been made as
typical use cases for cooperative perception (Ambrosin et al. 2019) with
obstructions from buildings or other vehicles on the open road with
other road-users. An overview of the experimental setup is shown at
Figure~\ref{fig-photo_dataset}.

The three experimental vehicles are equipped with a Velodyne VLP-32C
LiDAR mounted on their roof and a Novatel SPAN-CPT IMU with
Post-Processed Kinematics (PPK) corrections providing centimeter-level
localization accuracy. In addition, a Velodyne VLS-128 LiDAR sensor
statically placed on a curb was used as a roadside sensor. All sensors
are synchronized with GNSS time with a known extrinsic calibration.

\hypertarget{map-generation-using-real-data}{%
\subsection{Map Generation Using Real
Data}\label{map-generation-using-real-data}}

\begin{figure}

{\centering \includegraphics{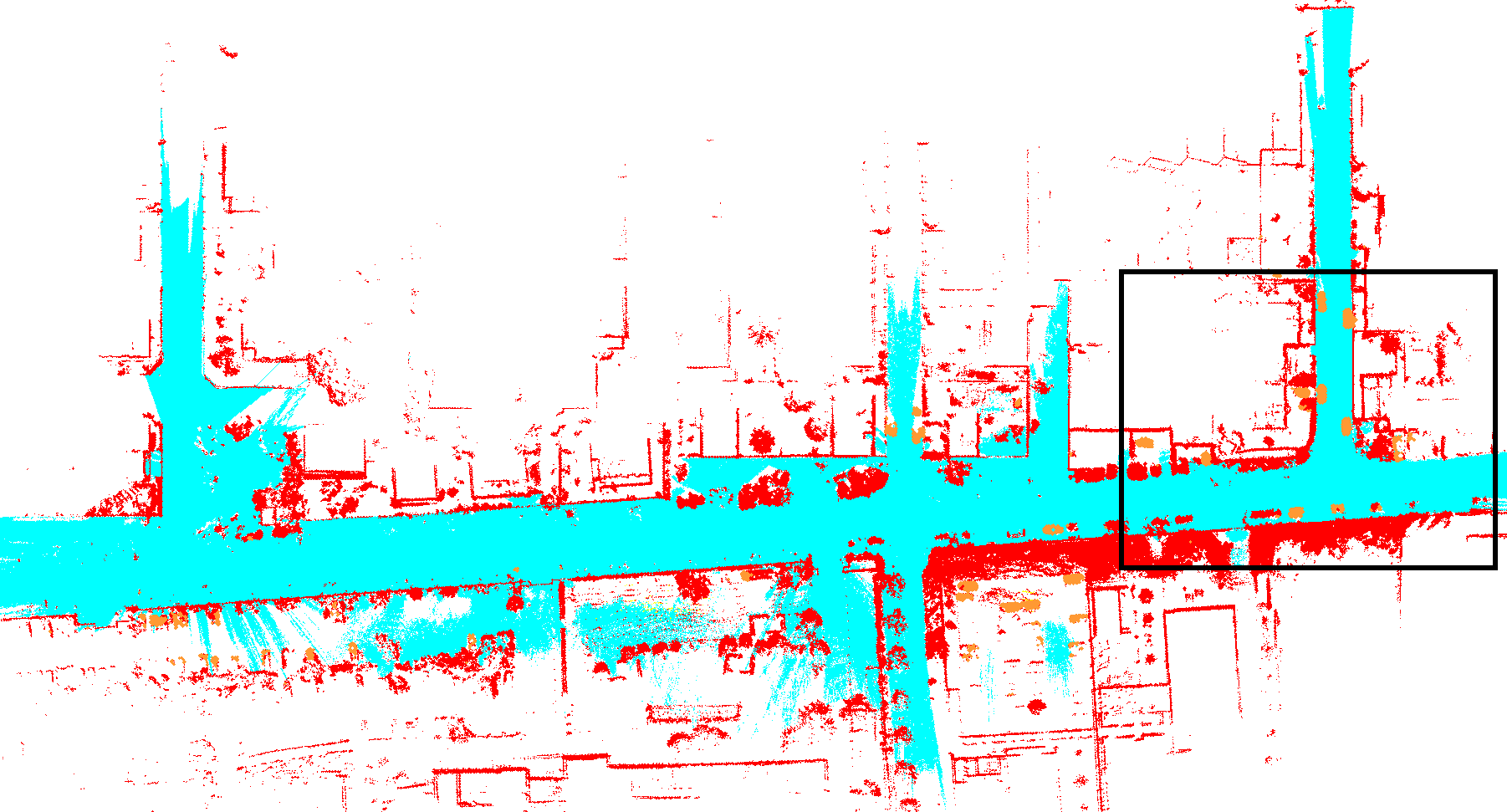}

}

\caption{\label{fig-generated_map}Generated map. Colors are the same as
described in Figure~\ref{fig-mapping}. The black rectangle is the area
zoomed-in for Figure~\ref{fig-map_zoom}.}

\end{figure}

\begin{figure}

{\centering \includegraphics{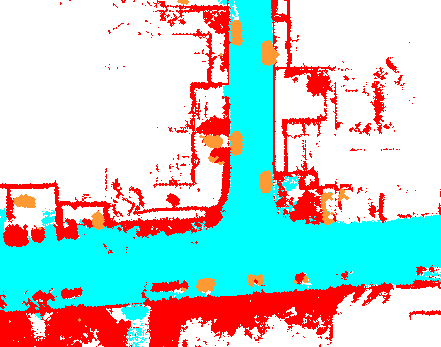}

}

\caption{\label{fig-map_zoom}Zoom on the top-right intersection of the
map. Uncertain zones (white), immovable occupancy (red), static objects
(orange) and passable space (cyan) are visible.}

\end{figure}

Accurate poses and point clouds have been thus recorded and processed
according to Section~\ref{sec-construction:map_generation}. In
particular, the threshold \(t_o\) and \(t_f\) were manually tuned to
\(5\) and \(30\), respectively (see Figure~\ref{fig-thresholds} for the
effects of the threshold tuning). The thresholds appear to be dependant
of the dataset, and can vary with the number of sensors or the number of
passages in a given area.

The map resulting from this processing is illustrated in
Figure~\ref{fig-generated_map}, Figure~\ref{fig-map_zoom}, respectively
showing the whole generated map and a zoom on a point of interest
(top-right intersection).

A proper evaluation of the map has been realized and shows interesting
results: passable areas fit well with the road, parked cars are well
identified as static and tree/buildings are well identified as
immovable.

\hypertarget{fusion-of-map-and-observations}{%
\subsection{Fusion of Map and
Observations}\label{fusion-of-map-and-observations}}

\begin{figure}

{\centering \includegraphics{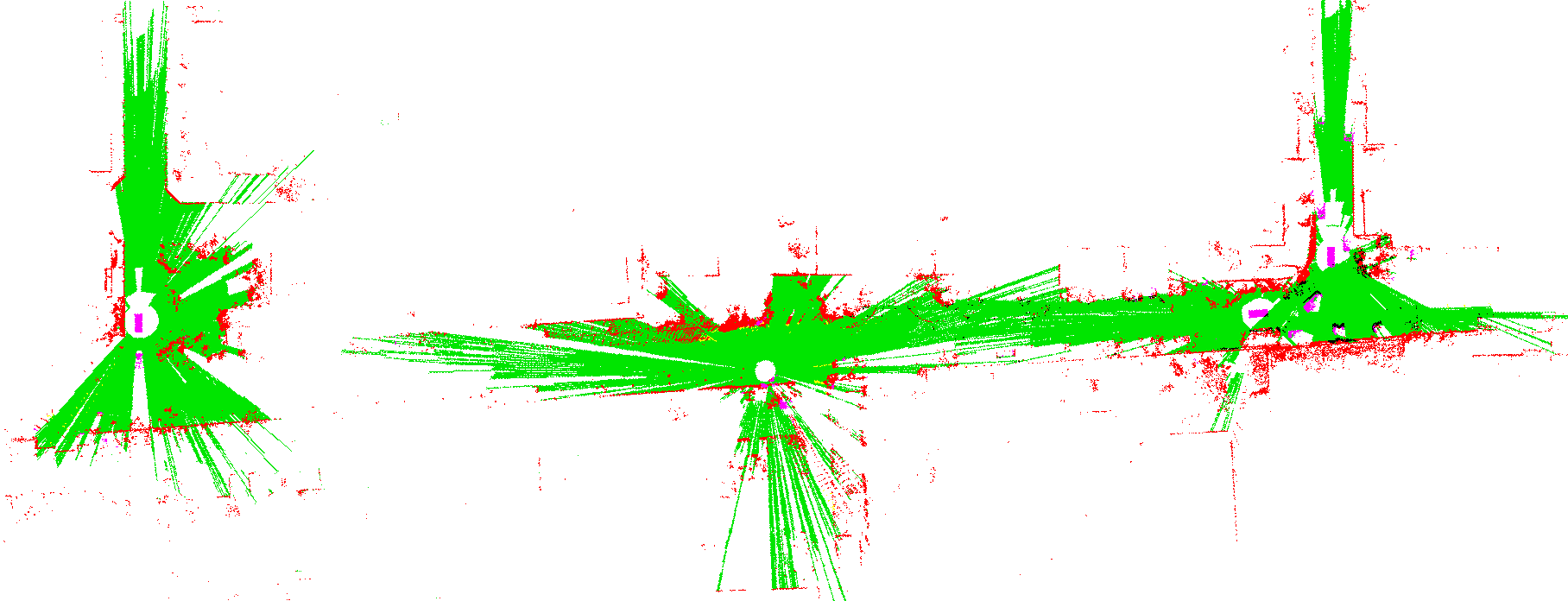}

}

\caption{\label{fig-observation}Fused observation grid. It contains all
the LiDAR observation at a given timestamp and the position of the
vehicles. Black cells represent conflicts between point of views, green
cells represent free space.}

\end{figure}

The observations contain the perception (LiDAR) from our vehicles and
their position and orientation as shown in Figure~\ref{fig-observation}.

\begin{figure}

{\centering \includegraphics{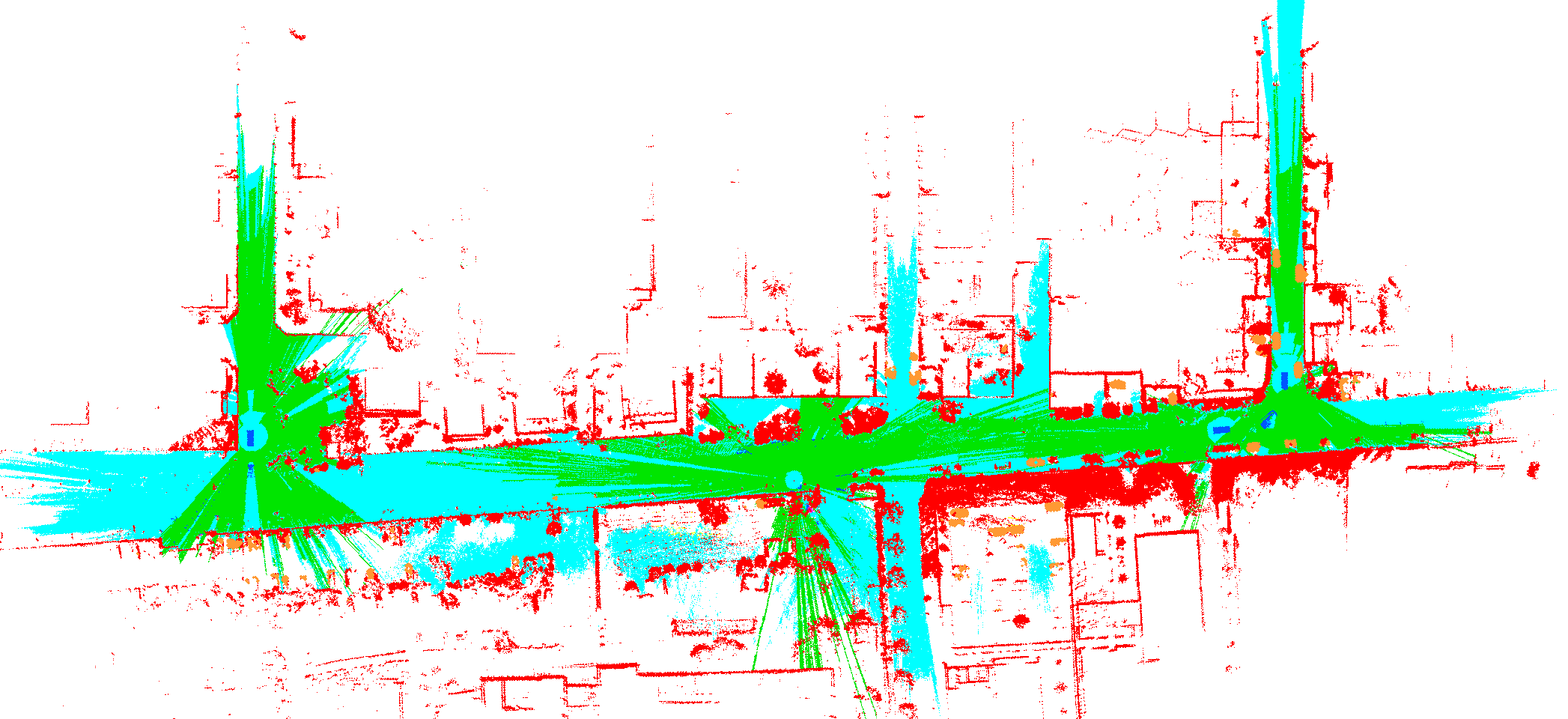}

}

\caption{\label{fig-fusion_map_obs}Fusion of map and observations. The
colors are the same as explained in Figure~\ref{fig-mapping}.}

\end{figure}

\begin{figure}

{\centering \includegraphics{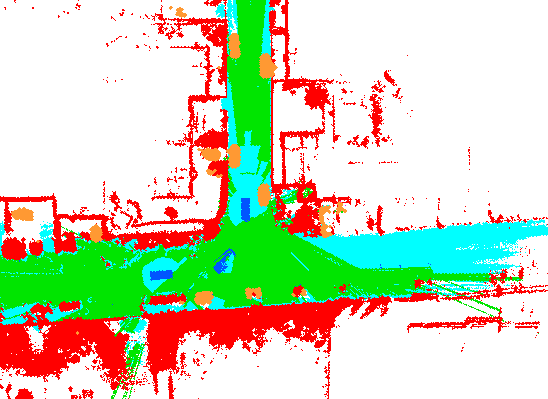}

}

\caption{\label{fig-fusion_map_obs_zoom}Zoom on
Figure~\ref{fig-fusion_map_obs}. Dynamic objects create blue cells
unless static evidence (orange cells) is present in the background map.}

\end{figure}

The result of the fusion is illustrated in
Figure~\ref{fig-fusion_map_obs} and Figure~\ref{fig-fusion_map_obs_zoom}
at a given time step and is evaluated in Section~\ref{sec-eval}.
Qualitatively, one can see that the free space is observed on top of the
passable zones of the map. Obstacles observed on top of passable zones
are correctly classified as dynamic objects. On the opposite, occupancy
observed over statically occupied zones correctly classifies them as
static \(S\) and thus prevents the creation of dynamic cells.

\hypertarget{sec-eval}{%
\subsection{Evaluation of the Method}\label{sec-eval}}

Existing datasets are not relevant to evaluate this method, as they
either lack moving vehicles or real data. A particular sequence of the
dataset introduced in Section~\ref{sec-dataset} has thus been manually
labeled to this end. It consists of 5 to 10 objects over 50 seconds or
500 frames. The manual labeling methodology is as follows: based on
accumulated point clouds, bird-eye-view bounding boxes have been
delimited by a human operator for several key frames of the dataset.
Bounding boxes have then been synchronized with the output of the above
method by interpolation. The center \(g_k^t=\left<x_k^t,y_k^t\right>\)
of each bounding box \(k\) for each time \(t\) has then been
concatenated to form the ground truth
\(\mathcal{G} = \left\{g_k^t\right\}_{t=1,2,\ldots}\). Finally, a Region
of Interest \(R\) is also delimited by the human operator. Inside \(R\),
the human operator certifies that no objects were missed, but objects
outside of it were ignored.

As our interest is about object existence and not spatial accuracy, we
evaluate our method object-wise. To do this, objects are first extracted
from the background and final maps from of
Section~\ref{sec-construction:map_generation} and
Section~\ref{sec-final_map}. The extraction is done by clustering static
cells in the background map and dynamic cells in final grids using the
DBSCAN algorithm (Ester et al. 1996). The center
\(o_j^t=\left<x_j^t,y_j^t\right>\) of each cluster \(j\) for each time
\(t\) are then concatenated to form the object list
\(\mathcal{O}=\left\{o_j^t\right\}_{t=1,2,\ldots}\). In a second time,
objects of \(\mathcal{O}\) that fall within \(R\) are compared to
\(\mathcal{G}\). Objects and ground truth are associated for each time
step using the Hungarian algorithm and Euclidean distances. Metrics are
then derived from these associations, with associated objects being True
Positives (\(TP\)), un-associated objects being False Positives (\(FP\))
and un-associated ground-truth being False Negatives (\(FN\)). Classical
metrics such as precision \(\left(\frac{TP}{TP+FP}\right)\), recall
\(\left(\frac{TP}{TP+FN}\right)\) and F1-score
\(\left(\frac{2TP}{2TP+FP+FN}\right)\) are provided in
Table~\ref{tbl-eval}. We believe that these scores are satisfactory as a
first pass to generate ground-truths, but that human verification still
remains mandatory.

\hypertarget{tbl-eval}{}
\begin{longtable}[]{@{}ccc@{}}
\caption{\label{tbl-eval}Evaluation of the proposed
method}\tabularnewline
\toprule\noalign{}
Precision & Recall & F1-score \\
\midrule\noalign{}
\endfirsthead
\toprule\noalign{}
Precision & Recall & F1-score \\
\midrule\noalign{}
\endhead
\bottomrule\noalign{}
\endlastfoot
\(0.95\) & \(0.97\) & \(0.96\) \\
\end{longtable}

\hypertarget{sec-conclusion}{%
\section{Conclusion and Future Work}\label{sec-conclusion}}

In this work, we proposed a novel approach to take advantage of
post-processing for ground truth semi-automatic generation by
introducing a complete frame of discernment for environment
representation and a \emph{background map} constructed using the whole
dataset. The fusion of the observation grids with the map allows the
instantaneous classification of objects as dynamic when they are on
passable areas, in contrast to other GTAM techniques (Steyer,
Tanzmeister, and Wollherr 2017; Tanzmeister and Wollherr 2017).
Moreover, we provided a dataset with multiple real moving points of
view, on which our method proved to have very good results.

Further improvements can be made such as tracking at object level above
the clustered dynamic cells or the generation of parametric free space
above the grid. We believe that tracking the dynamic objects using
filtering and smoothing algorithms would lead to better results as it
would remove some noise from the detection and would help to follow
dynamic obstacles in occluded environments. Additionally, the current
implementation was made using binary masses. A further task would be to
implement finer masses using confidence provided by the sensors or
processing methods. This would allow a finer representation of
uncertainty during the fusion of different information sources.

\hypertarget{acknowledgment}{%
\section*{Acknowledgment}\label{acknowledgment}}
\addcontentsline{toc}{section}{Acknowledgment}

This work has been carried out within the SIVALab laboratory between
Renault and Heudiasyc and co-funded by the Région Hauts-de-France. The
authors would like to thank C. Zinoune for lending the VLS-128 LiDAR and
the colleagues of the lab who participated to the acquisition of the
dataset, in particular S. Bonnet.

\hypertarget{references}{%
\section*{References}\label{references}}
\addcontentsline{toc}{section}{References}

\hypertarget{refs}{}
\begin{CSLReferences}{1}{0}
\leavevmode\vadjust pre{\hypertarget{ref-ambrosin_object-level_2019}{}}%
Ambrosin, Moreno, Ignacio J Alvarez, Cornelius Buerkle, Lily L Yang,
Fabian Oboril, Manoj R Sastry, and Kathiravetpillai Sivanesan. 2019.
{``Object-Level {Perception Sharing Among Connected Vehicles}.''} In
\emph{{IEEE Intelligent Transportation Systems Conference}}.

\leavevmode\vadjust pre{\hypertarget{ref-behleySemanticKITTIDatasetSemantic2019}{}}%
Behley, J., M. Garbade, A. Milioto, J. Quenzel, S. Behnke, C. Stachniss,
and J. Gall. 2019. {``{SemanticKITTI: A Dataset for Semantic Scene
Understanding of LiDAR Sequences}.''} In \emph{Proc. Of the IEEE/CVF
International Conf.~On Computer Vision (ICCV)}.

\leavevmode\vadjust pre{\hypertarget{ref-buschLUMPILeibnizUniversity2022}{}}%
Busch, Steffen, Christian Koetsier, Jeldrik Axmann, and Claus Brenner.
2022. {``LUMPI: The Leibniz University Multi-Perspective Intersection
Dataset.''} In \emph{2022 IEEE Intelligent Vehicles Symposium (IV)},
1127--34. \url{https://doi.org/10.1109/IV51971.2022.9827157}.

\leavevmode\vadjust pre{\hypertarget{ref-caesarNuScenesMultimodalDataset2020}{}}%
Caesar, Holger, Varun Bankiti, Alex H. Lang, Sourabh Vora, Venice Erin
Liong, Qiang Xu, Anush Krishnan, Yu Pan, Giancarlo Baldan, and Oscar
Beijbom. 2020. {``nuScenes: A Multimodal Dataset for Autonomous
Driving.''} In \emph{CVPR}.

\leavevmode\vadjust pre{\hypertarget{ref-chenMovingObjectSegmentation2021}{}}%
Chen, Xieyuanli, Shijie Li, Benedikt Mersch, Louis Wiesmann, Jurgen
Gall, Jens Behley, and Cyrill Stachniss. 2021. {``Moving {Object
Segmentation} in {3D LiDAR Data}: {A Learning-Based Approach Exploiting
Sequential Data}.''} \emph{IEEE Robotics and Automation Letters}.

\leavevmode\vadjust pre{\hypertarget{ref-coueBayesianOccupancyFiltering2006}{}}%
Coué, Christophe, Cédric Pradalier, Christian Laugier, Thierry
Fraichard, and Pierre Bessière. 2006. {``Bayesian {Occupancy Filtering}
for {Multitarget Tracking}: {An Automotive Application}.''} \emph{The
International Journal of Robotics Research}.

\leavevmode\vadjust pre{\hypertarget{ref-dempsterGeneralizationBayesianInference1968}{}}%
Dempster, A. P. 1968. {``A {Generalization} of {Bayesian Inference}.''}
\emph{Journal of the Royal Statistical Society. Series B
(Methodological)}.

\leavevmode\vadjust pre{\hypertarget{ref-dewanMotionbasedDetectionTracking2016}{}}%
Dewan, Ayush, Tim Caselitz, Gian Diego Tipaldi, and Wolfram Burgard.
2016. {``Motion-Based Detection and Tracking in {3D LiDAR} Scans.''} In
\emph{{IEEE International Conference} on {Robotics} and {Automation}}.

\leavevmode\vadjust pre{\hypertarget{ref-elfesUsingOccupancyGrids1989}{}}%
Elfes, A. 1989. {``Using Occupancy Grids for Mobile Robot Perception and
Navigation.''} \emph{Computer}.

\leavevmode\vadjust pre{\hypertarget{ref-erikstelletPostProcessingLaser2016}{}}%
Erik Stellet, Jan, Leopold Walkling, and J. Marius Zöllner. 2016.
{``Post Processing of Laser Scanner Measurements for Testing Advanced
Driver Assistance Systems.''} In \emph{{International Conference} on
{Information Fusion}}.

\leavevmode\vadjust pre{\hypertarget{ref-ester_density-based_1996}{}}%
Ester, Martin, Hans-Peter Kriegel, Jörg Sander, and Xiaowei Xu. 1996.
{``A Density-Based Algorithm for Discovering Clusters in Large Spatial
Databases with Noise.''} In \emph{International {Conference} on
{Knowledge Discovery} and {Data Mining}}.

\leavevmode\vadjust pre{\hypertarget{ref-Ettinger_2021_ICCV}{}}%
Ettinger, Scott, Shuyang Cheng, Benjamin Caine, Chenxi Liu, Hang Zhao,
Sabeek Pradhan, Yuning Chai, et al. 2021. {``Large Scale Interactive
Motion Forecasting for Autonomous Driving: The Waymo Open Motion
Dataset.''} In \emph{IEEE/CVF International Conference on Computer
Vision}.

\leavevmode\vadjust pre{\hypertarget{ref-geigerAreWeReady2012}{}}%
Geiger, A., P. Lenz, and R. Urtasun. 2012. {``Are We Ready for
Autonomous Driving? {The KITTI} Vision Benchmark Suite.''} In
\emph{{IEEE Conference} on {Computer Vision} and {Pattern Recognition}}.

\leavevmode\vadjust pre{\hypertarget{ref-hanCollaborativePerceptionAutonomous2023}{}}%
Han, Yushan, Hui Zhang, Huifang Li, Yi Jin, Congyan Lang, and Yidong Li.
2023. {``Collaborative Perception in Autonomous Driving: Methods,
Datasets and Challenges.''} \emph{ArXiv} abs/2301.06262.

\leavevmode\vadjust pre{\hypertarget{ref-jimenezGroundSegmentationAlgorithm2021}{}}%
Jiménez, Víctor, Jorge Godoy, Antonio Artuñedo, and Jorge Villagra.
2021. {``Ground {Segmentation Algorithm} for {Sloped Terrain} and
{Sparse LiDAR Point Cloud}.''} \emph{IEEE Access}.

\leavevmode\vadjust pre{\hypertarget{ref-liV2XSimMultiAgentCollaborative2022}{}}%
Li, Yiming, Dekun Ma, Ziyan An, Zixun Wang, Yiqi Zhong, Siheng Chen, and
Chen Feng. 2022. {``V2X-Sim: Multi-Agent Collaborative Perception
Dataset and Benchmark for Autonomous Driving.''} \emph{IEEE Robotics and
Automation Letters} 7 (4): 10914--21.
\url{https://doi.org/10.1109/LRA.2022.3192802}.

\leavevmode\vadjust pre{\hypertarget{ref-liDeepLearningLiDAR2020}{}}%
Li, Ying, Lingfei Ma, Zilong Zhong, Fei Liu, Dongpu Cao, Jonathan Li,
and Michael A. Chapman. 2020. {``Deep Learning for LiDAR Point Clouds in
Autonomous Driving: A Review.''} \emph{IEEE Transactions on Neural
Networks and Learning Systems} 32: 3412--32.

\leavevmode\vadjust pre{\hypertarget{ref-maoDOLPHINSDatasetCollaborative2022}{}}%
Mao, Ruiqing, Jingyu Guo, Yukuan Jia, Yuxuan Sun, Sheng Zhou, and
Zhisheng Niu. 2022. {``{DOLPHINS}: {Dataset} for {Collaborative
Perception} Enabled {Harmonious} and {Interconnected Self-driving}.''}
In \emph{Asian Conference on Computer Vision}.

\leavevmode\vadjust pre{\hypertarget{ref-moras_credibilist_2011}{}}%
Moras, Julien, Véronique Cherfaoui, and Philippe Bonnifait. 2011.
{``Credibilist Occupancy Grids for Vehicle Perception in Dynamic
Environments.''} In \emph{{IEEE International Conference} on {Robotics}
and {Automation}}.

\leavevmode\vadjust pre{\hypertarget{ref-shaferMathematicalTheoryEvidence1976}{}}%
Shafer, Glenn. 1976. \emph{A {Mathematical Theory} of {Evidence}}.
{Princeton University Press}.

\leavevmode\vadjust pre{\hypertarget{ref-steyerObjectTrackingBased2017}{}}%
Steyer, Sascha, Georg Tanzmeister, and Dirk Wollherr. 2017. {``Object
Tracking Based on Evidential Dynamic Occupancy Grids in Urban
Environments.''} In \emph{{IEEE Intelligent Vehicles Symposium} ({IV})}.

\leavevmode\vadjust pre{\hypertarget{ref-steyerGridBasedEnvironmentEstimation2018}{}}%
---------. 2018. {``Grid-{Based Environment Estimation Using Evidential
Mapping} and {Particle Tracking}.''} \emph{IEEE Transactions on
Intelligent Vehicles}.

\leavevmode\vadjust pre{\hypertarget{ref-tanzmeisterEvidentialGridBasedTracking2017}{}}%
Tanzmeister, Georg, and Dirk Wollherr. 2017. {``Evidential {Grid-Based
Tracking} and {Mapping}.''} \emph{IEEE Transactions on Intelligent
Transportation Systems}.

\leavevmode\vadjust pre{\hypertarget{ref-voMultitargetTracking2015}{}}%
Vo, Ba-ngu, Mahendra Mallick, Yaakov Bar-shalom, Stefano Coraluppi,
Richard Osborne, Ronald Mahler, and Ba-tuong Vo. 2015. {``Multitarget
{Tracking}.''} In \emph{Wiley {Encyclopedia} of {Electrical} and
{Electronics Engineering}}. {John Wiley \& Sons, Inc.}

\leavevmode\vadjust pre{\hypertarget{ref-xu_opv2v_2022}{}}%
Xu, Runsheng, Hao Xiang, Xin Xia, Xu Han, Jinlong Li, and Jiaqi Ma.
2022. {``{OPV2V}: {An Open Benchmark Dataset} and {Fusion Pipeline} for
{Perception} with {Vehicle-to-Vehicle Communication}.''} In
\emph{{International Conference} on {Robotics} and {Automation}}.

\leavevmode\vadjust pre{\hypertarget{ref-yeCooperativeRawSensor2020}{}}%
Ye, Egon, Philip Spiegel, and Matthias Althoff. 2020. {``Cooperative
{Raw Sensor Data Fusion} for {Ground Truth Generation} in {Autonomous
Driving}.''} In \emph{{IEEE} {International Conference} on {Intelligent
Transportation Systems}}.

\leavevmode\vadjust pre{\hypertarget{ref-yeOfflineDynamicGrid2021}{}}%
Ye, Egon, Gerald Wursching, Sascha Steyer, and Matthias Althoff. 2021.
{``Offline {Dynamic Grid Generation} for {Automotive Environment
Perception Using Temporal Inference Methods}.''} \emph{IEEE Robotics and
Automation Letters}.

\leavevmode\vadjust pre{\hypertarget{ref-yuTrackBeforeDetectLabeledMultiBernoulli2020}{}}%
Yu, Boqian, and Egon Ye. 2020. {``Track-{Before-Detect Labeled
Multi-Bernoulli Smoothing} for {Multiple Extended Objects}.''} In
\emph{{IEEE} {International Conference} on {Information Fusion}}.

\leavevmode\vadjust pre{\hypertarget{ref-yuDAIRV2XLargeScaleDataset2022}{}}%
Yu, Haibao, Yizhen Luo, Mao Shu, Yiyi Huo, Zebang Yang, Yifeng Shi,
Zhenglong Guo, et al. 2022. {``{DAIR-V2X}: {A Large-Scale Dataset} for
{Vehicle-Infrastructure Cooperative 3D Object Detection}.''} In
\emph{{IEEE}/{CVF Conference} on {Computer Vision} and {Pattern
Recognition}}.

\leavevmode\vadjust pre{\hypertarget{ref-zamanakosComprehensiveSurveyLIDARbased2021}{}}%
Zamanakos, Georgios, Lazaros Tsochatzidis, Angelos Amanatiadis, and
Ioannis Pratikakis. 2021. {``A Comprehensive Survey of {LIDAR-based 3D}
Object Detection Methods with Deep Learning for Autonomous Driving.''}
\emph{Computers \& Graphics}.

\leavevmode\vadjust pre{\hypertarget{ref-zhuCylindricalAsymmetrical3D2021}{}}%
Zhu, Xinge, Hui Zhou, Tai Wang, Fangzhou Hong, Yuexin Ma, Wei Li,
Hongsheng Li, and Dahua Lin. 2021. {``Cylindrical and {Asymmetrical 3D
Convolution Networks} for {LiDAR Segmentation}.''} In \emph{{IEEE}/{CVF
Conference} on {Computer Vision} and {Pattern Recognition}}.

\leavevmode\vadjust pre{\hypertarget{ref-ziegler_fast_2010}{}}%
Ziegler, Julius, and Christoph Stiller. 2010. {``Fast Collision Checking
for Intelligent Vehicle Motion Planning.''} In \emph{IEEE Intelligent
Vehicles Symposium}.

\end{CSLReferences}

\end{document}